\documentclass[11pt,a4paper]{article}
\usepackage[hyperref]{acl2020}
\usepackage{times}
\usepackage{latexsym}

\usepackage{microtype}
\usepackage{graphicx}
\usepackage{amsmath}
\usepackage{amsthm}
\usepackage{amsfonts}
\usepackage{amssymb}
\usepackage{bbm}
\usepackage{booktabs}
\usepackage{colortbl}
\usepackage{xspace}
\usepackage{color}
\usepackage{multirow}
\usepackage{comment}
\usepackage{pifont}
\definecolor{deepblue}{rgb}{0,0,0.5}
\definecolor{deepred}{rgb}{0.6,0,0}
\definecolor{deepgreen}{rgb}{0,0.5,0}
\definecolor{darkgreen}{RGB}{43,163,39}
\definecolor{bluesquare}{rgb}{126,166,224}

\usepackage{graphicx}
\usepackage{bm}

\def\tightcol{\hskip 6pt}

\usepackage{listings}

\lstdefinestyle{pythoncode}{
  language=Python,
  morekeywords={self,join,append,split,write,chain,sub,lower,strip,decode,choice,listdir,savefig,expanduser,translate},             
  keywordstyle=\bfseries\color{deepblue},
  emph={zip,decode,translate,savefig,compile,expanduser,isoformat},          
  emphstyle=\color{deepred},    
  showstringspaces=false,
  breaklines=true,
  escapeinside=||,
  columns=fullflexible,
  basicstyle=\fontfamily{cmtt}\small,
  belowskip=0.2\baselineskip,
  aboveskip=-0.6\baselineskip
}

\newcommand{\inlineCode}{\lstinline[mathescape,style=pythoncode]}

\newcolumntype{H}{>{\setbox0=\hbox\bgroup}c<{\egroup}@{}}

\newcommand{\refmark}{\ding{51}\xspace}
\newcommand{\cmark}{$\clubsuit$\xspace}
\newcommand{\xmark}{$\spadesuit$\xspace}

\newcommand*\ttvar[1]{\texttt{\expandafter\dottvar\detokenize{#1}\relax}}
\newcommand*\dottvar[1]{\ifx\relax#1\else
  \expandafter\ifx\string(#1\string(\allowbreak\else#1\fi
  \expandafter\dottvar\fi}

\aclfinalcopy 
\def\aclpaperid{576} 

\title{Incorporating External Knowledge through Pre-training \\ for Natural Language to Code Generation}

\author{
Frank F. Xu\thanks{~~The first two authors contributed equally.}, Zhengbao Jiang\footnotemark[1], Pengcheng Yin, Bogdan Vasilescu, Graham Neubig \\
Carnegie Mellon University \\
\texttt{\{fangzhex,zhengbaj,pcyin,vasilescu,gneubig\}@cs.cmu.edu}
}

\date{}

\begin{document}
\maketitle
\begin{abstract}
Open-domain code generation aims to generate code in a general-purpose programming language (such as Python) from natural language (NL) intents.
Motivated by the intuition that developers usually retrieve resources on the web when writing code, 
we explore the effectiveness of incorporating two varieties of external knowledge into NL-to-code generation: automatically mined NL-code pairs from the online programming QA forum StackOverflow and programming language API documentation.
Our evaluations show that combining the two sources with data augmentation and retrieval-based data re-sampling improves the current state-of-the-art by up to 2.2\% absolute BLEU score on the code generation testbed CoNaLa.
The code and resources are available at \url{https://github.com/neulab/external-knowledge-codegen}.
\end{abstract}

\section{Introduction}





Semantic parsing, the task of generating machine executable meaning 
representations from natural language (NL) intents, has generally focused 
on limited domains~\cite{zelle1996learning,data-atis-original}, or 
domain-specific languages with a limited set of operators~\cite{berant-etal-2013-semantic,quirk-etal-2015-language,dong-lapata-2016-language,liang-etal-2017-neural,krishnamurthy-etal-2017-neural,zhong2017seq2sql,yu-etal-2018-spider,yu-etal-2019-sparc,yu-etal-2019-cosql}. 
However, recently there has been a move towards applying semantic 
parsing to automatically generating source code in general-purpose programming languages~\cite{yin2018mining,yao2018staqc,lin-etal-2018-nl2bash,agashe-etal-2019-juice,yao2019coacor}. 
Prior work in this area~\cite{xiao-etal-2016-sequence,ling-etal-2016-latent,rabinovich-etal-2017-abstract,yin-neubig-2017-syntactic,yin-neubig-2018-tranx,dong-lapata-2018-coarse,
suhr-etal-2018-learning,iyer-etal-2018-mapping,yin-neubig-2019-reranking} used a variety of models,
especially neural architectures, to achieve good performance.

\begin{figure}[t]
    \centering
    \includegraphics[width=0.9\linewidth]{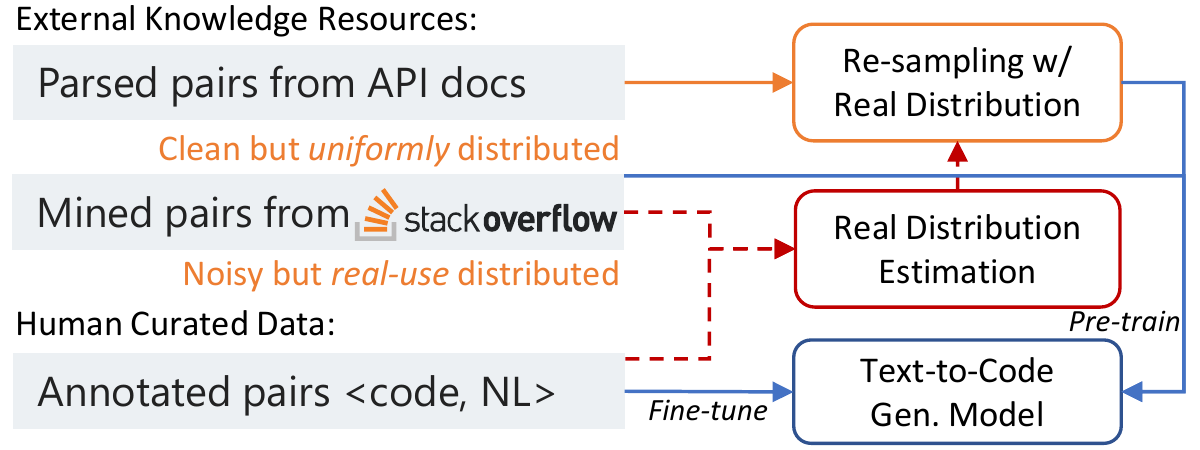}
    \vspace{-1mm}
    \caption{Our approach: incorporating external knowledge by data re-sampling, pre-training and fine-tuning.}
    \vspace{-2mm}
    \label{fig:overview}
\end{figure}

\looseness=-1
However, open-domain code generation for general-purpose languages 
like Python is challenging.
For example, given the intent to \textit{choose a random file from the 
directory contents of the C drive, `C:\textbackslash\textbackslash'}, 
one would expect the Python code snippet 
{\small\texttt{random.choice(os.listdir(`C:\textbackslash\textbackslash'))}}, 
that realizes the given intent.
This would involve not just generating syntactically correct code, but 
also using (and potentially combining) calls to APIs and libraries that 
implement some of the desired functionality.
As we show in \autoref{sec:exp}, current code generation models 
still have difficulty generating the correct function calls with 
appropriate argument placement.
For example, given the NL intent above, although the state-of-the-art model by 
\citet{yin-neubig-2018-tranx} that uses a transition-based method to generate Python abstract syntax trees is guaranteed to generate \textit{syntactically correct} code, it still incorrectly outputs 
{\small\ttvar{random.savefig(random(compile(open(`C:\\'))+100).isoformat())}}.

\looseness=-1
A known bottleneck to training more accurate code generation models is 
the limited number of manually annotated training pairs available in 
existing human-curated datasets, which are insufficient to cover the myriad of ways in which some complex functionality could be implemented in code.
However, increasing the size of labeled datasets through additional human
annotation is relatively expensive.
It is also the case that human developers rarely reference such paired examples
of NL and code, and rather take external resources on the web and modify them into the desired form~\cite{brandt2009two,brandt2010example,gu2016deep}.
Motivated by these facts, we propose to improve the performance of code 
generation models through a novel training strategy:
pre-training the model on data extracted automatically from external 
knowledge resources such as existing API documentation, 
before fine-tuning it on
a small manually curated dataset (\autoref{sec:pre-training}).
Our approach, outlined in \autoref{fig:overview}, combines pairs of NL intents and code snippets mined 
automatically from the Q\&A website StackOverflow (\autoref{sec:mined}), 
and API documentation for common software libraries (\autoref{sec:api}).\footnote{
Of course external knowledge for code covers a large variety of resources, other than these two types.}

\looseness=-1
While our approach is model-agnostic and generally applicable, we implement
it on top of a state-of-the-art syntax-based method for code 
generation, TranX~\cite{yin-neubig-2018-tranx}, with additional 
hypothesis reranking~\cite{yin-neubig-2019-reranking}.
Experiments on the CoNaLa benchmark~\cite{yin2018mining} show that 
incorporating external knowledge through our proposed methods increases 
BLEU score from 30.1 to 32.3, outperforming the previous state-of-the-art 
model by up to 2.2\% absolute.
Qualitatively analyzing a sample of code snippets generated by our model
reveals that the generated code is more likely to use the correct API calls 
for desired functionality and to arrange arguments in the right order.

\section{Approach}
\label{sec:data_sources}

\subsection{Over-arching Framework}
\label{sec:pre-training}

The overall strategy for incorporating external knowledge that we take on this work is to (1) \emph{pre-train} the model on the NL-code pairs obtained from external resources, then (2) \emph{fine-tune} on a small manually curated corpus.
This allows the model to first learn on larger amounts of potentially noisy data, while finally being tailored to the actual NL and code we want to model at test time.
In order to perform this pre-training we need to convert external data sources into NL-code pairs, and we describe how to do so in the following sections.

\subsection{Mined NL-code Pairs}
\label{sec:mined}
When developers code, most will inevitably search online for code snippets demonstrating how to achieve their particular intent.
One of the most prominent resources online is StackOverflow,\footnote{\url{https://stackoverflow.com}} a popular programming QA forum.
However, it is not the case that all code on StackOverflow actually reflects the corresponding intent stated by the questioner -- some may be methods defining variables or importing necessary libraries, while other code may be completely irrelevant.
\citet{yin2018mining} propose training a classifier to decide whether an NL-code pair is valid, resulting in a large but noisy parallel corpus of NL intents and source code snippets.
The probability assigned by the method can serve as confidence, representing the quality of the automatically mined NL-code pairs.
We use these mined pairs as a first source of external knowledge.


\subsection{API Documentation}
\label{sec:api}
Second, motivated by the intuition that much of modern software development relies on libraries, and that developers often turn to programming language and software library references for help while writing code, 
we consider API documentation as another source of external knowledge.

\begin{figure}
\centering
\includegraphics[width=0.9\columnwidth]{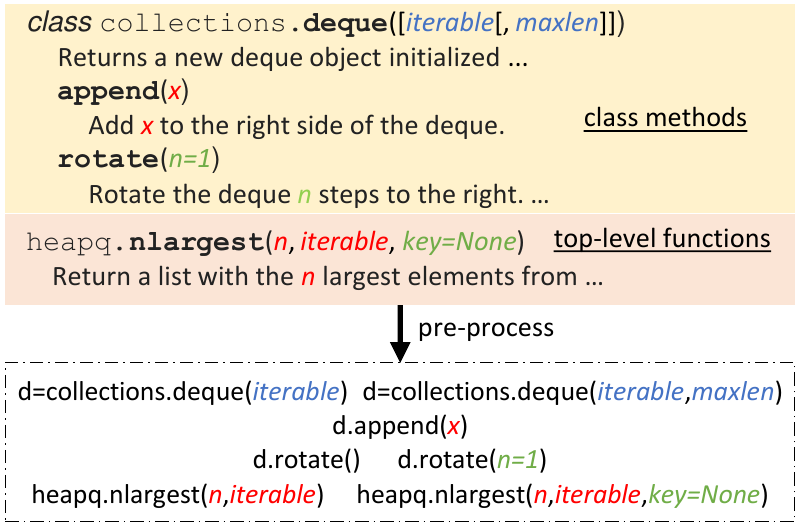}
\caption{Examples from Python API documentation and pre-processed code snippets, including class constructors, methods, and top-level functions.
We use \textcolor{red}{red}, \textcolor{blue}{blue}, and \textcolor{green}{green} to denote required, optional positional, and optional keyword arguments respectively.}
\vspace{-2mm}
\label{fig:apidoc}
\end{figure}

\autoref{fig:apidoc} shows some examples from the Python standard library API documentation.
It contains descriptions of libraries, classes, methods, functions, and arguments.
The documentation is already in a paired form consisting of code signatures and their descriptions. 
However, the signatures shown in the documentation mainly provide the prototype of the API rather than valid API usages appearing in source code.
The text descriptions in the documentation tend to be verbose for clarity, while real questions from developers are usually succinct.
We use a few heuristics to transform these to emulate real inputs a code generation system may face.

Most APIs define required and optional arguments in the signature.
In real usage, developers usually provide none or only some of those arguments.
To simulate this, we permute all possible combinations (with a limit) of the optional arguments and append them to the required arguments, following correct syntax.
For class constructors and methods, we create a heuristic variable name based on the class name to store the instantiated class object and to call methods upon.
To make concise description for each code snippet created, we preserve only the first sentence in the corresponding documentation, as well as the first sentences that contain mentions of each argument in the snippet.
In the rare case where arguments are not found in the original description, we add another sentence containing these arguments to the end of the NL snippet, ensuring all variables in code are covered in the NL.
We detail this process in Appendix~\ref{app:apidocs}.

\subsection{Re-sampling API Knowledge}
\label{sec:re-sampling}


External knowledge from different sources has different characteristics.
NL-code pairs automatically mined from StackOverflow are good representatives of the questions that developers may ask, but are inevitably noisy.
NL-code pairs from API documentation are clean, but there may be a topical distribution shift from real questions asked by developers.
For example, the library \texttt{curses} has significantly more API entries than \texttt{json} (178 vs. 17),\footnote{\url{https://docs.python.org/3.7/library/curses.html} and \url{https://docs.python.org/3.7/library/json.html}} while \texttt{json} is more frequently asked about and used.
This distributional shift between pre-training and fine-tuning causes performance degradation, as shown later in \autoref{sec:exp_result}.



To mitigate this problem, we propose a retrieval-based re-sampling method to close the gap between the API documentation and the actual NL-code pairs we want to model.
We use both human annotated data $\mathcal{D}_\text{ann}$ and mined data $\mathcal{D}_\text{mine}$ to model the distribution of NL-code pairs because they are both produced by real users.
For each sample in this real usage distribution, we retrieve $k$ NL-code pairs from the set of pairs harvested from API documentation $\mathcal{D}_\text{API}$ and aggregate the frequencies of each pair $y \in \mathcal{D}_\text{API}$ being retrieved:
\begin{equation*}
\text{freq}(y) = \sum_{x \in \mathcal{D}_\text{ann+mined}} \delta(y \in R(x, \mathcal{D}_\text{API}, k)),
\end{equation*}
where $R(x, \mathcal{D}_\text{API}, k)$ retrieves the top $k$ most similar samples from $\mathcal{D}_\text{API}$ given $x$, either according to NL intent or code snippet.
$\delta(\cdot)$ is Kronecker's delta function, returning 1 if the internal condition is true, and 0 otherwise. 
We use the BM25 retrieval algorithm \cite{SPARCKJONES2000779} implemented in ElasticSearch.\footnote{\url{https://github.com/elastic/elasticsearch}. When retrieving with code snippets, all the punctuation marks are removed.}
We take this frequency and calculate the probability distribution after smoothing with a temperature $\tau \in [1,\infty]$:
\begin{equation*}
    \small
    P(y) = \text{freq}(y)^{1/\tau} / \sum_{y' \in \mathcal{D}_\text{API}} \text{freq}(y')^{1/\tau}
\end{equation*}
As $\tau$ changes from 1 to $\infty$, $P(y)$ shifts from a distribution proportional to the frequency to a uniform distribution.
Using this distribution, we can sample NL-code pairs from the API documentation that are more likely to be widely-used API calls.

\section{Experiments}
\label{sec:exp}

\subsection{Experimental Settings}
\noindent \textbf{Dataset and Metric:}
Although the proposed approach is generally applicable and model-agnostic, for evaluation purposes, we choose CoNaLa~\cite{yin2018mining} as the human-annotated dataset (2,179 training, 200 dev and 500 test samples). 
It covers real-world English queries about Python with diverse intents.
We use the same evaluation metric as the CoNaLa benchmark, corpus-level BLEU calculated on target code outputs in test set.

\noindent \textbf{Mined Pairs:}
We use the CoNaLa-Mined~\cite{yin2018mining} dataset of 600K NL-code pairs in Python automatically mined from StackOverflow (\autoref{sec:mined}).
We sort all pairs by their confidence scores, and found that approximately top 100K samples are of reasonable quality in terms of code correctness and NL-code correspondence.
We therefore choose the top 100K pairs for the experiments.

\noindent \textbf{API Documentation Pairs:} 
We parsed all the module documentation including libraries, built-in types and functions included in the Python 3.7.5 distribution.\footnote{\url{https://docs.python.org/release/3.7.5/library/index.html}}
After pre-processing (\autoref{sec:api}), we create about 13K distinct NL-code pairs (without re-sampling) from Python API documentation.
For fair comparison, we also sample the same number of pairs for the re-sampling setting (\autoref{sec:re-sampling}).

\noindent \textbf{Methods:}
We choose the current state-of-the-art NL-to-code generation model TranX~\cite{yin-neubig-2018-tranx} with hypothesis reranking~\cite{yin-neubig-2019-reranking} as the base model.
Plus, we incorporate length normalization~\cite{cho-2014-learning} to prevent beam search from favoring shorter results over longer ones.
\textbf{Man} denotes training solely on CoNaLa.
\textbf{Man+Mine} refers to first pre-training on mined data, then fine-tuning on CoNaLa.
\textbf{Man+Mine+API} combines both mined data and API documentation for pre-training.
As a comparison to our distribution-based method (denoted by \textbf{dist.}, \autoref{sec:re-sampling}), we also attempt to directly retrieve top 5 NL-code pairs from API documents (denoted by \textbf{direct}).\footnote{We choose 5 to obtain comparable amount of pairs.}

\noindent \textbf{Implementation Details:}
We experiment with $k=\{1,3,5\}$ and $\tau=\{1,2,5\}$ in re-sampling, and find that $k=1$ and $\tau=2$ perform the best.
We follow the original hyper-parameters in TranX, except that we use a batch size of 64 and 10 in pre-training and fine-tuning respectively.

\subsection{Results}\label{sec:exp_result}

\begin{table}[t]
\begin{center}
\small
\begin{tabular}{lllr}
\toprule
\textbf{Data Strategy} & \multicolumn{2}{l}{\textbf{Method}} & \textbf{BLEU} \\
\midrule
Man & & & 27.20 \\
\midrule
\multirow{2}{*}{Man+Mine}& \multicolumn{2}{l}{50k} & 27.94 \\
& \multicolumn{2}{l}{100k} & 28.14 \\
\midrule
\multirow{5}{*}{Man+Mine+API} & \multicolumn{2}{c}{w/o re-sampling} & 27.84 \\
& direct & intent & 29.66 \\
& dist. & intent & 29.31 \\
& direct & code & 30.26 \\
& dist. & code & \textbf{30.69} \\
\midrule
Man & \multicolumn{2}{l}{\multirow{3}{*}{+rerank}} & 30.11  \\
Man+Mine(100k) & \multicolumn{2}{c}{} & 31.42 \\
Our best & \multicolumn{2}{c}{} & \textbf{32.26} \\
\bottomrule
\end{tabular}
\end{center}
\vspace{-2.5mm}
\caption{Performance comparison of different strategies to incorporate external knowledge.}
\vspace{-3mm}
\label{tab:result}
\end{table}

Results are summarized in \autoref{tab:result}.
We can first see that by incorporating more noisy mined data during pre-training allows for a small improvement due to increased coverage from the much larger training set.
Further, if we add the pairs harvested from API docs for pre-training without re-sampling the performance drops, validating the challenge of distributional shift mentioned in \autoref{sec:re-sampling}.

Comparing the two re-sampling strategies \textbf{direct} vs. \textbf{dist.}, and two different retrieval targets NL intent vs. code snippet, we can see that \textbf{dist.} performs better with the code snippet as the retrieval target.
We expect that using code snippets to retrieve pairs performs better because it makes the generation \emph{target}, the code snippet, more similar to the real-world distribution, thus better training the decoder.
It is also partly because API descriptions are inherently different than questions asked by developers (e.g.~they have more verbose wording), causing intent retrieval to be less accurate. 

Lastly, we apply hypothesis reranking to both the base model and our best approach and find improvements afforded by our proposed strategy of incorporating external knowledge are mostly orthogonal to those from hypothesis reranking.

After showing the effectiveness of our proposed re-sampling strategy, we are interested in the performance on more-used versus less-used APIs for the potentially skewed overall performance.
We use string matching heuristics to obtain the standard Python APIs used in the dataset and calculated the average frequency of API usages in each data instance.
We then select the top 200 and the bottom 200 instances out of the 500 test samples in terms of API usage frequencies. 
Before and after adding API docs into pre-training, the BLEU score on both splits saw improvements: for high-frequency split, it goes from 28.67 to 30.91 and for low-frequency split, it goes from 27.55 to 30.05, indicating that although the re-sampling would skew towards high-frequency APIs, with the appropriate smoothing temperature experimentation, it will still contribute to performance increases on low-frequency APIs.

Besides using BLEU scores to perform holistic evaluation, we also perform more fine-grained analysis of what types of tokens generated are improving.
We apply heuristics on the abstract syntax tree of the generated code to identify tokens for API calls and variable names in the test data, and calculated the token-level accuracy for each. 
The API call accuracy increases from 31.5\% to 36.8\% and the variable name accuracy from 41.2\% to 43.0\% after adding external resources, meaning that both the API calls and argument usages are getting better using our approach.


\begin{table}[tb]
\begin{center}
\small
\begin{tabular}{l@{\tightcol}p{7.0cm}}
\toprule
\multicolumn{2}{l}{Open a file ``\texttt{f.txt}'' in write mode.} \\
\refmark & \inlineCode+f=open(`f.txt', `w')+  \\
\xmark & \inlineCode+f=open(`f.txt', $\text{\color{deepred}`f.txt'}$)+  \\
\cmark & \inlineCode+f=open(`f.txt', `w')+  \\
\midrule
\multicolumn{2}{p{7.2cm}}{lower a string \textit{text} and remove non-alphanumeric characters aside from space.} \\
\refmark & \inlineCode+re.sub(r`[^\sa-zA-Z0-9]', `', text).lower().strip()+ \\
\xmark & \inlineCode+text.decode.translate(text.strip(), $\text{\color{deepred}`non-alphanumeric'}$, `')+  \\
\cmark & \inlineCode+re.sub(r`[^\sa-zA-Z0-9]', `', text)+ \\
\midrule
\multicolumn{2}{p{7.2cm}}{choose a random file from the directory contents of the C drive, `C:\textbackslash\textbackslash'.} \\
\refmark & \inlineCode+random.choice(os.listdir(`C:\\'))+ \\
\xmark & \inlineCode-random.savefig($\text{\color{deepred}random}$(compile($\text{\color{deepred}open}$(`C:\\'))+100).$\text{\color{deepred}isoformat()}$)-  \\
\cmark & \inlineCode+random.choice(os.path.expanduser(`C:\\'))+\\
\bottomrule
\end{tabular}
\end{center}
\vspace{-2.5mm}
\caption{Examples, where
\refmark is the ground-truth code snippet, \xmark is the original output, and \cmark is the output with our proposed methods.
Correct and erroneous function calls are marked in {\color{deepblue}blue} and {\color{deepred}red} respectively.}
\vspace{-3mm}
\label{tab:case}
\end{table}

\subsection{Case Study}
We further show selected outputs from both the baseline and our best approach in~\autoref{tab:case}.
In general, we can see that the NL to code generation task is still challenging, especially with more complex intents that require nested or chained API calls, or functions with more arguments.
The vanilla model already can generate basic functions and copy strings/variables to the output, but we observe that incorporating external knowledge improves the results in two main ways: 
1) better argument placement for APIs, and 2) better selection of which API call should be used for a certain intent.

In the first example, we can see that although the baseline gets the function call ``\texttt{open()}'' correct, it fails to generate the correct second argument specifying write mode, while our approach is able to successfully generate the appropriate \texttt{`w'}.
In the second and third example, we can see that the baseline uses the wrong API calls, and sometimes ``makes up'' APIs on its own (e.g. ``\texttt{random.savefig()}'').
However, our approach's outputs, while not perfect, are much more successful at generating correct API calls that actually exist and make sense for the intent.

On a closer look, we can observe that both the addition of mined examples and API docs may have brought the improvement.
The example of the ``\texttt{open()}'' function added from API docs uses the default mode ``\texttt{r}'', so learning the meaning of ``\texttt{w}'' argument is due to the added mined real examples, but learning the argument placement (first file name as a string, second a shorthand mode identifier as a character) may have occurred from the API docs. 
In other examples, ``\texttt{random.choice()}'' and ``\texttt{re.sub()}'' both are Python standard library APIs so they are included in the API doc examples.

\section{Conclusion and Future Work}
We proposed a model-agnostic approach based on data augmentation, retrieval and data re-sampling, to incorporate external knowledge into code generation models, which achieved state-of-the-art results on the CoNaLa open-domain code generation task.

In the future, evaluation by automatically executing generated code with test cases could be a better way to assess code generation results. 
It will also likely be useful to generalize our re-sampling procedures to zero-shot scenarios, where a programmer writes a library and documents it, but nobody has used it yet.
For example, developers may provide relative estimates of each documented API usages to guide the re-sampling; or we could find nearest neighbors to each API call in terms of semantics and use existing usage statistics as estimates to guide the re-sampling. 

\section*{Acknowledgments}

This research was supported by NSF Award No. 1815287 ``Open-domain, Data-driven Code Synthesis from Natural Language.''

\bibliography{anthology,acl2020}

\begin{thebibliography}{30}
\expandafter\ifx\csname natexlab\endcsname\relax\def\natexlab#1{#1}\fi

\bibitem[{Agashe et~al.(2019)Agashe, Iyer, and
  Zettlemoyer}]{agashe-etal-2019-juice}
Rajas Agashe, Srinivasan Iyer, and Luke Zettlemoyer. 2019.
\newblock \href {https://doi.org/10.18653/v1/D19-1546} {{J}u{IC}e: A large
  scale distantly supervised dataset for open domain context-based code
  generation}.
\newblock In \emph{Proceedings of the 2019 Conference on Empirical Methods in
  Natural Language Processing and the 9th International Joint Conference on
  Natural Language Processing (EMNLP-IJCNLP)}, pages 5435--5445, Hong Kong,
  China. Association for Computational Linguistics.

\bibitem[{Berant et~al.(2013)Berant, Chou, Frostig, and
  Liang}]{berant-etal-2013-semantic}
Jonathan Berant, Andrew Chou, Roy Frostig, and Percy Liang. 2013.
\newblock \href {https://www.aclweb.org/anthology/D13-1160} {Semantic parsing
  on {F}reebase from question-answer pairs}.
\newblock In \emph{Proceedings of the 2013 Conference on Empirical Methods in
  Natural Language Processing}, pages 1533--1544, Seattle, Washington, USA.
  Association for Computational Linguistics.

\bibitem[{Brandt et~al.(2010)Brandt, Dontcheva, Weskamp, and
  Klemmer}]{brandt2010example}
Joel Brandt, Mira Dontcheva, Marcos Weskamp, and Scott~R Klemmer. 2010.
\newblock \href
  {https://hci.stanford.edu/publications/2010/blueprint/brandt_chi10_blueprint.pdf}
  {Example-centric programming: integrating web search into the development
  environment}.
\newblock In \emph{Proceedings of the SIGCHI Conference on Human Factors in
  Computing Systems}, pages 513--522. ACM.

\bibitem[{Brandt et~al.(2009)Brandt, Guo, Lewenstein, Dontcheva, and
  Klemmer}]{brandt2009two}
Joel Brandt, Philip~J Guo, Joel Lewenstein, Mira Dontcheva, and Scott~R
  Klemmer. 2009.
\newblock \href
  {http://pgbovine.net/publications/opportunistic-programming-two-studies_CHI-2009.pdf}
  {Two studies of opportunistic programming: interleaving web foraging,
  learning, and writing code}.
\newblock In \emph{Proceedings of the SIGCHI Conference on Human Factors in
  Computing Systems}, pages 1589--1598. ACM.

\bibitem[{Cho et~al.(2014)Cho, van Merri{\"e}nboer, Gulcehre, Bahdanau,
  Bougares, Schwenk, and Bengio}]{cho-2014-learning}
Kyunghyun Cho, Bart van Merri{\"e}nboer, Caglar Gulcehre, Dzmitry Bahdanau,
  Fethi Bougares, Holger Schwenk, and Yoshua Bengio. 2014.
\newblock \href {https://doi.org/10.3115/v1/D14-1179} {Learning phrase
  representations using {RNN} encoder{--}decoder for statistical machine
  translation}.
\newblock In \emph{Proceedings of the 2014 Conference on Empirical Methods in
  Natural Language Processing ({EMNLP})}, pages 1724--1734, Doha, Qatar.
  Association for Computational Linguistics.

\bibitem[{Deborah A.~Dahl and Shriber(1994)}]{data-atis-original}
Michael Brown William Fisher Kate Hunicke-Smith David Pallett Christine Pao
  Alexander~Rudnicky Deborah A.~Dahl, Madeleine~Bates and Elizabeth Shriber.
  1994.
\newblock \href {http://dl.acm.org/citation.cfm?id=1075823} {{Expanding the
  scope of the ATIS task: The ATIS-3 corpus}}.
\newblock \emph{Proceedings of the workshop on Human Language Technology},
  pages 43--48.

\bibitem[{Dong and Lapata(2016)}]{dong-lapata-2016-language}
Li~Dong and Mirella Lapata. 2016.
\newblock \href {https://doi.org/10.18653/v1/P16-1004} {Language to logical
  form with neural attention}.
\newblock In \emph{Proceedings of the 54th Annual Meeting of the Association
  for Computational Linguistics (Volume 1: Long Papers)}, pages 33--43, Berlin,
  Germany. Association for Computational Linguistics.

\bibitem[{Dong and Lapata(2018)}]{dong-lapata-2018-coarse}
Li~Dong and Mirella Lapata. 2018.
\newblock \href {https://doi.org/10.18653/v1/P18-1068} {Coarse-to-fine decoding
  for neural semantic parsing}.
\newblock In \emph{Proceedings of the 56th Annual Meeting of the Association
  for Computational Linguistics (Volume 1: Long Papers)}, pages 731--742,
  Melbourne, Australia. Association for Computational Linguistics.

\bibitem[{Gu et~al.(2016)Gu, Zhang, Zhang, and Kim}]{gu2016deep}
Xiaodong Gu, Hongyu Zhang, Dongmei Zhang, and Sunghun Kim. 2016.
\newblock \href {https://dl.acm.org/citation.cfm?id=2950334} {Deep {API}
  learning}.
\newblock In \emph{Proceedings of the 2016 24th ACM SIGSOFT International
  Symposium on Foundations of Software Engineering}, pages 631--642. ACM.

\bibitem[{Iyer et~al.(2018)Iyer, Konstas, Cheung, and
  Zettlemoyer}]{iyer-etal-2018-mapping}
Srinivasan Iyer, Ioannis Konstas, Alvin Cheung, and Luke Zettlemoyer. 2018.
\newblock \href {https://doi.org/10.18653/v1/D18-1192} {Mapping language to
  code in programmatic context}.
\newblock In \emph{Proceedings of the 2018 Conference on Empirical Methods in
  Natural Language Processing}, pages 1643--1652, Brussels, Belgium.
  Association for Computational Linguistics.

\bibitem[{Jones et~al.(2000)Jones, Walker, and Robertson}]{SPARCKJONES2000779}
K.~Sparck Jones, S.~Walker, and S.E. Robertson. 2000.
\newblock \href {https://doi.org/https://doi.org/10.1016/S0306-4573(00)00015-7}
  {A probabilistic model of information retrieval: development and comparative
  experiments: Part 1}.
\newblock \emph{Information Processing \& Management}, 36(6):779 -- 808.

\bibitem[{Krishnamurthy et~al.(2017)Krishnamurthy, Dasigi, and
  Gardner}]{krishnamurthy-etal-2017-neural}
Jayant Krishnamurthy, Pradeep Dasigi, and Matt Gardner. 2017.
\newblock \href {https://doi.org/10.18653/v1/D17-1160} {Neural semantic parsing
  with type constraints for semi-structured tables}.
\newblock In \emph{Proceedings of the 2017 Conference on Empirical Methods in
  Natural Language Processing}, pages 1516--1526, Copenhagen, Denmark.
  Association for Computational Linguistics.

\bibitem[{Liang et~al.(2017)Liang, Berant, Le, Forbus, and
  Lao}]{liang-etal-2017-neural}
Chen Liang, Jonathan Berant, Quoc Le, Kenneth~D. Forbus, and Ni~Lao. 2017.
\newblock \href {https://doi.org/10.18653/v1/P17-1003} {Neural symbolic
  machines: Learning semantic parsers on {F}reebase with weak supervision}.
\newblock In \emph{Proceedings of the 55th Annual Meeting of the Association
  for Computational Linguistics (Volume 1: Long Papers)}, pages 23--33,
  Vancouver, Canada. Association for Computational Linguistics.

\bibitem[{Lin et~al.(2018)Lin, Wang, Zettlemoyer, and
  Ernst}]{lin-etal-2018-nl2bash}
Xi~Victoria Lin, Chenglong Wang, Luke Zettlemoyer, and Michael~D. Ernst. 2018.
\newblock \href {https://www.aclweb.org/anthology/L18-1491} {{NL}2{B}ash: A
  corpus and semantic parser for natural language interface to the linux
  operating system}.
\newblock In \emph{Proceedings of the Eleventh International Conference on
  Language Resources and Evaluation ({LREC}-2018)}, Miyazaki, Japan. European
  Languages Resources Association (ELRA).

\bibitem[{Ling et~al.(2016)Ling, Blunsom, Grefenstette, Hermann,
  Ko{\v{c}}isk{\'y}, Wang, and Senior}]{ling-etal-2016-latent}
Wang Ling, Phil Blunsom, Edward Grefenstette, Karl~Moritz Hermann,
  Tom{\'a}{\v{s}} Ko{\v{c}}isk{\'y}, Fumin Wang, and Andrew Senior. 2016.
\newblock \href {https://doi.org/10.18653/v1/P16-1057} {Latent predictor
  networks for code generation}.
\newblock In \emph{Proceedings of the 54th Annual Meeting of the Association
  for Computational Linguistics (Volume 1: Long Papers)}, pages 599--609,
  Berlin, Germany. Association for Computational Linguistics.

\bibitem[{Quirk et~al.(2015)Quirk, Mooney, and
  Galley}]{quirk-etal-2015-language}
Chris Quirk, Raymond Mooney, and Michel Galley. 2015.
\newblock \href {https://doi.org/10.3115/v1/P15-1085} {Language to code:
  Learning semantic parsers for if-this-then-that recipes}.
\newblock In \emph{Proceedings of the 53rd Annual Meeting of the Association
  for Computational Linguistics and the 7th International Joint Conference on
  Natural Language Processing (Volume 1: Long Papers)}, pages 878--888,
  Beijing, China. Association for Computational Linguistics.

\bibitem[{Rabinovich et~al.(2017)Rabinovich, Stern, and
  Klein}]{rabinovich-etal-2017-abstract}
Maxim Rabinovich, Mitchell Stern, and Dan Klein. 2017.
\newblock \href {https://doi.org/10.18653/v1/P17-1105} {Abstract syntax
  networks for code generation and semantic parsing}.
\newblock In \emph{Proceedings of the 55th Annual Meeting of the Association
  for Computational Linguistics (Volume 1: Long Papers)}, pages 1139--1149,
  Vancouver, Canada. Association for Computational Linguistics.

\bibitem[{Suhr et~al.(2018)Suhr, Iyer, and Artzi}]{suhr-etal-2018-learning}
Alane Suhr, Srinivasan Iyer, and Yoav Artzi. 2018.
\newblock \href {https://doi.org/10.18653/v1/N18-1203} {Learning to map
  context-dependent sentences to executable formal queries}.
\newblock In \emph{Proceedings of the 2018 Conference of the North {A}merican
  Chapter of the Association for Computational Linguistics: Human Language
  Technologies, Volume 1 (Long Papers)}, pages 2238--2249, New Orleans,
  Louisiana. Association for Computational Linguistics.

\bibitem[{Xiao et~al.(2016)Xiao, Dymetman, and
  Gardent}]{xiao-etal-2016-sequence}
Chunyang Xiao, Marc Dymetman, and Claire Gardent. 2016.
\newblock \href {https://doi.org/10.18653/v1/P16-1127} {Sequence-based
  structured prediction for semantic parsing}.
\newblock In \emph{Proceedings of the 54th Annual Meeting of the Association
  for Computational Linguistics (Volume 1: Long Papers)}, pages 1341--1350,
  Berlin, Germany. Association for Computational Linguistics.

\bibitem[{Yao et~al.(2019)Yao, Peddamail, and Sun}]{yao2019coacor}
Ziyu Yao, Jayavardhan~Reddy Peddamail, and Huan Sun. 2019.
\newblock \href {https://dl.acm.org/citation.cfm?id=3313632} {Coacor: Code
  annotation for code retrieval with reinforcement learning}.
\newblock In \emph{The World Wide Web Conference}, pages 2203--2214. ACM.

\bibitem[{Yao et~al.(2018)Yao, Weld, Chen, and Sun}]{yao2018staqc}
Ziyu Yao, Daniel~S Weld, Wei-Peng Chen, and Huan Sun. 2018.
\newblock \href {https://dl.acm.org/citation.cfm?id=3186081} {Staqc: A
  systematically mined question-code dataset from stack overflow}.
\newblock In \emph{Proceedings of the 2018 World Wide Web Conference}, pages
  1693--1703. International World Wide Web Conferences Steering Committee.

\bibitem[{Yin et~al.(2018)Yin, Deng, Chen, Vasilescu, and
  Neubig}]{yin2018mining}
Pengcheng Yin, Bowen Deng, Edgar Chen, Bogdan Vasilescu, and Graham Neubig.
  2018.
\newblock \href {https://doi.org/https://doi.org/10.1145/3196398.3196408}
  {Learning to mine aligned code and natural language pairs from stack
  overflow}.
\newblock In \emph{International Conference on Mining Software Repositories},
  MSR, pages 476--486. ACM.

\bibitem[{Yin and Neubig(2017)}]{yin-neubig-2017-syntactic}
Pengcheng Yin and Graham Neubig. 2017.
\newblock \href {https://doi.org/10.18653/v1/P17-1041} {A syntactic neural
  model for general-purpose code generation}.
\newblock In \emph{Proceedings of the 55th Annual Meeting of the Association
  for Computational Linguistics (Volume 1: Long Papers)}, pages 440--450,
  Vancouver, Canada. Association for Computational Linguistics.

\bibitem[{Yin and Neubig(2018)}]{yin-neubig-2018-tranx}
Pengcheng Yin and Graham Neubig. 2018.
\newblock \href {https://doi.org/10.18653/v1/D18-2002} {{TRANX}: A
  transition-based neural abstract syntax parser for semantic parsing and code
  generation}.
\newblock In \emph{Proceedings of the 2018 Conference on Empirical Methods in
  Natural Language Processing: System Demonstrations}, pages 7--12, Brussels,
  Belgium. Association for Computational Linguistics.

\bibitem[{Yin and Neubig(2019)}]{yin-neubig-2019-reranking}
Pengcheng Yin and Graham Neubig. 2019.
\newblock \href {https://doi.org/10.18653/v1/P19-1447} {Reranking for neural
  semantic parsing}.
\newblock In \emph{Proceedings of the 57th Annual Meeting of the Association
  for Computational Linguistics}, pages 4553--4559, Florence, Italy.
  Association for Computational Linguistics.

\bibitem[{Yu et~al.(2019{\natexlab{a}})Yu, Zhang, Er, Li, Xue, Pang, Lin, Tan,
  Shi, Li, Jiang, Yasunaga, Shim, Chen, Fabbri, Li, Chen, Zhang, Dixit, Zhang,
  Xiong, Socher, Lasecki, and Radev}]{yu-etal-2019-cosql}
Tao Yu, Rui Zhang, Heyang Er, Suyi Li, Eric Xue, Bo~Pang, Xi~Victoria Lin,
  Yi~Chern Tan, Tianze Shi, Zihan Li, Youxuan Jiang, Michihiro Yasunaga,
  Sungrok Shim, Tao Chen, Alexander Fabbri, Zifan Li, Luyao Chen, Yuwen Zhang,
  Shreya Dixit, Vincent Zhang, Caiming Xiong, Richard Socher, Walter Lasecki,
  and Dragomir Radev. 2019{\natexlab{a}}.
\newblock \href {https://doi.org/10.18653/v1/D19-1204} {{C}o{SQL}: A
  conversational text-to-{SQL} challenge towards cross-domain natural language
  interfaces to databases}.
\newblock In \emph{Proceedings of the 2019 Conference on Empirical Methods in
  Natural Language Processing and the 9th International Joint Conference on
  Natural Language Processing (EMNLP-IJCNLP)}, pages 1962--1979, Hong Kong,
  China. Association for Computational Linguistics.

\bibitem[{Yu et~al.(2018)Yu, Zhang, Yang, Yasunaga, Wang, Li, Ma, Li, Yao,
  Roman, Zhang, and Radev}]{yu-etal-2018-spider}
Tao Yu, Rui Zhang, Kai Yang, Michihiro Yasunaga, Dongxu Wang, Zifan Li, James
  Ma, Irene Li, Qingning Yao, Shanelle Roman, Zilin Zhang, and Dragomir Radev.
  2018.
\newblock \href {https://doi.org/10.18653/v1/D18-1425} {{S}pider: A large-scale
  human-labeled dataset for complex and cross-domain semantic parsing and
  text-to-{SQL} task}.
\newblock In \emph{Proceedings of the 2018 Conference on Empirical Methods in
  Natural Language Processing}, pages 3911--3921, Brussels, Belgium.
  Association for Computational Linguistics.

\bibitem[{Yu et~al.(2019{\natexlab{b}})Yu, Zhang, Yasunaga, Tan, Lin, Li, Er,
  Li, Pang, Chen, Ji, Dixit, Proctor, Shim, Kraft, Zhang, Xiong, Socher, and
  Radev}]{yu-etal-2019-sparc}
Tao Yu, Rui Zhang, Michihiro Yasunaga, Yi~Chern Tan, Xi~Victoria Lin, Suyi Li,
  Heyang Er, Irene Li, Bo~Pang, Tao Chen, Emily Ji, Shreya Dixit, David
  Proctor, Sungrok Shim, Jonathan Kraft, Vincent Zhang, Caiming Xiong, Richard
  Socher, and Dragomir Radev. 2019{\natexlab{b}}.
\newblock \href {https://doi.org/10.18653/v1/P19-1443} {{SP}ar{C}: Cross-domain
  semantic parsing in context}.
\newblock In \emph{Proceedings of the 57th Annual Meeting of the Association
  for Computational Linguistics}, pages 4511--4523, Florence, Italy.
  Association for Computational Linguistics.

\bibitem[{Zelle and Mooney(1996)}]{zelle1996learning}
John~M Zelle and Raymond~J Mooney. 1996.
\newblock \href {http://dl.acm.org/citation.cfm?id=1075823} {Learning to parse
  database queries using inductive logic programming}.
\newblock In \emph{Proceedings of the national conference on artificial
  intelligence}, pages 1050--1055.

\bibitem[{Zhong et~al.(2017)Zhong, Xiong, and Socher}]{zhong2017seq2sql}
Victor Zhong, Caiming Xiong, and Richard Socher. 2017.
\newblock \href {https://arxiv.org/abs/1709.00103} {Seq2sql: Generating
  structured queries from natural language using reinforcement learning}.
\newblock \emph{arXiv preprint arXiv:1709.00103}.

\end{thebibliography}
\bibliographystyle{acl_natbib}

\newpage
\appendix

\section{API Documentation Pre-processing}
\label{app:apidocs}
Here we describe detailed heuristics used for API documentation preprocessing. 
The goal is to harvest NL-code pairs with API docs as a source.
\subsection{Arguments}
Most APIs will have arguments, either required or optional.
For the required arguments, we leave them ``as-is''.
We deal with two types of optional arguments, positional arguments and keyword arguments through permutation and sampling. 
In the Python documentation, optional positional arguments are bracketed in ``\texttt{[.., [..]]}''.
Nested brackets are commonly used to represent more than one possible optional positional arguments.
Another type of optional arguments are implemented using keyword arguments in the form of \texttt{key=default}.

In real usage, developers usually only provide none or some of those arguments.
To simulate this, we permute all possible combinations of the optional arguments, and append them to the required arguments.
For example, if the code signature in the documentation writes ``\texttt{collections.deque([iterable[, maxlen]])}'', we produce all 3 possible usages: ``\texttt{collections.deque()}'', ``\texttt{collections.deque(iterable)}'', and ``\texttt{collections.deque(iterable, maxlen)}''.
For keyword arguments like ``\texttt{heapq.nlargest(n, iterable, key=None)}'',
we will also include ``\texttt{heapq.nlargest(n, iterable)}'' in addition.
The total number of permutations is $n+1$ for a function with $n$ optional positional arguments, and $2^n = \binom n0 + \binom n1 + ... + \binom nn$ for a function with $n$ optional keyword arguments, which leads to exponentially large number of samples for functions with many optional keywords.
Motivated by the observation that developers rarely specify all of the optional arguments, but rather tend to use default values, we only keep the top 10 permutations with the least number of optional arguments.

\subsection{Class Initializers and Methods} 
Other heuristics are used to transform code signatures related to classes to emulate real usage.
For class initializers in the documentation, we construct an assignment statement with lower-cased variable name using the first character of the class name to store the instantiated class, e.g. \texttt{d = collections.deque(iterable)}.
For class methods, we prepend a heuristically created variable name to the method call, emulating a real method call on an instantiated class, e.g. \texttt{d.append(x)}.

\subsection{Documentation} 
Official documentation tends to be verbose for clarity, while real questions from developers are usually succinct.
Thus we use the following heuristics to keep only sentences in the document that are necessary for generating the code as the intent text.
We include the first sentence because it usually describes the functionality of the API.
For each argument in the emulated API usage code snippet, we include the first sentence in the documentation that mentions the argument through string matching.
For arguments not mentioned in the documentation, we add a sentence in the end: ``With arguments \texttt{'arg\_name'} ...'' to ensure all arguments are covered verbatim in the intent text.

\end{document}